\pdfoutput=1

\documentclass[11pt]{article}

\usepackage[preprint]{acl}

\usepackage{times}
\usepackage{latexsym}
\usepackage{microtype}
\usepackage{hyperref}
\usepackage{url}
\usepackage{booktabs}
\usepackage{subcaption}
\definecolor{darkblue}{rgb}{0, 0, 0.5}
\hypersetup{colorlinks=true, citecolor=darkblue, linkcolor=darkblue, urlcolor=darkblue}

\usepackage{algorithm}
\usepackage{makecell}
\usepackage{color}
\usepackage{amssymb}
\usepackage{amsmath}
\usepackage{bbding}
\usepackage{algpseudocode}
\usepackage{algorithmicx,algorithm}

\usepackage{multirow}
\usepackage{booktabs}
\usepackage{graphicx}

\newcommand{\guanhua}[1]{{\color{black} #1}}

\usepackage[T1]{fontenc}

\usepackage[utf8]{inputenc}

\usepackage{microtype}

\usepackage{graphicx}

\title{Not All LoRA Parameters Are Essential: Insights on Inference Necessity}

\author{
 \textbf{Guanhua Chen\textsuperscript{1}},
 \textbf{Yutong Yao\textsuperscript{1}},
 \textbf{Ci-Jun Gao\textsuperscript{2}},
 \textbf{Lidia S. Chao\textsuperscript{1}},
 \\
 \textbf{Feng Wan\textsuperscript{2}},
 \textbf{Derek F. Wong$^\dagger$\thanks{$^\dagger$ Corresponding author.}\textsuperscript{1}},
\\
 \textsuperscript{1}NLP\textsuperscript{2}CT Lab, Department of Computer and Information Science, University of Macau
 \\
 \textsuperscript{2}Department of Electrical and Computer Engineering, University of Macau,
\\
 \{nlp2ct.guanhua, nlp2ct.yutong, cijun.gao\}@gmail.com\\
         \{derekfw, lidiasc, fwan\}@um.edu.mo\\
}

\begin{document}
\maketitle
\begin{abstract}
Current research on LoRA primarily focuses on minimizing the number of fine-tuned parameters or optimizing its architecture. However, the necessity of all fine-tuned LoRA layers during inference remains underexplored. In this paper, we investigate the contribution of each LoRA layer to the model's ability to predict the ground truth and hypothesize that lower-layer LoRA modules play a more critical role in model reasoning and understanding. To address this, we propose a simple yet effective method to enhance the performance of large language models (LLMs) fine-tuned with LoRA. Specifically, we identify a ``boundary layer'' that distinguishes essential LoRA layers by analyzing a small set of validation samples. During inference, we drop all LoRA layers beyond this boundary. We evaluate our approach on three strong baselines across four widely-used text generation datasets. Our results demonstrate consistent and significant improvements, underscoring the effectiveness of selectively retaining critical LoRA layers during inference.

\end{abstract}

\section{Introduction}
Large language models (LLMs), such as ChatGPT, have demonstrated remarkable capabilities across diverse downstream tasks. However, their extensive parameterization presents substantial challenges for fine-tuning. In response, parameter-efficient fine-tuning (PEFT) methods \cite{houlsby2019parameter, li2021prefix} have gained significant traction. Notably, Low-Rank Adaptation (LoRA) has emerged as a pivotal technique, particularly in the context of LLMs. LoRA operates by introducing trainable adapters for each layer of the LLM while keeping the remaining parameters frozen. This approach not only substantially reduces the computational resources required for fine-tuning but also achieves performance that is on par with or even superior to fully fine-tuned LLMs.

To further leverage LoRA for improving training efficiency and model performance, various studies have focused on optimizing its architecture or pruning important parameters for each layer. \citet{zhang2023adaptive} introduced AdaLoRA, which utilizes singular value decomposition (SVD) of $\triangle W$ to dynamically adjust the rank of LoRA for different layers. LoRA-Drop \cite{zhou2024lora} prunes LoRA parameters based on output evaluation. HydraLoRA \cite{tian2024hydralora} proposes an asymmetric structure that employs a shared A matrix and multiple B matrices to handle complex domain datasets. MoELoRA \cite{luo2024moelora} selects suitable A and B matrices in LoRA for each layer to improve adaptation. However, these methods either focus on more efficient parameter fine-tuning or necessitate the fine-tuning of more complex LoRA structures. We argue that selectively using the fine-tuned LoRA of part of the layers without additional training will be more efficient, given the already small size of LoRA parameters.

In this work, we conduct a systematic investigation into the layer-wise impact of LoRA (Low-Rank Adaptation) in LLMs. Our empirical analysis reveals a distinct functional separation across model layers: the lower layers predominantly engage in content understanding and information extraction, while the upper layers specialize in answer summarization and refinement. Interestingly, our findings demonstrate that the top layers can effectively generate responses based on the representations captured by the bottom layers even without LoRA adaptation, leveraging the inherent knowledge encoded in the pre-trained LLMs. Building upon these observations, we propose a novel and computationally efficient approach to optimize LoRA-based fine-tuning. Specifically, we introduce two complementary strategies for identifying a critical ``boundary layer'' that separates information extraction from answer refinement in LoRA-enhanced models. The first strategy employs a manual approach by computing the average probability of ground truth outputs across LoRA-tuned layers using validation samples, with the boundary determined through probability curve analysis. The second strategy adopts an automated approach by evaluating model performance across different boundary layer configurations and selecting the optimal one. During inference, we strategically remove LoRA layers above the identified boundary layer, achieving both efficiency gains and performance maintenance.

We evaluate our proposed method on four widely-used generation datasets spanning multiple tasks, employing three state-of-the-art baselines: \textbf{Phi-2} \cite{textbooks2}, \textbf{Llama2-7B-Chat} \cite{touvron2023llama}, and \textbf{Llama-3.1-8B-Instruct} \cite{dubey2024llama}. Our empirical results consistently demonstrate the effectiveness and generaliz ability of the proposed approach across different model architectures and task domains.

\section{Related Work}
\subsection{Parameter-Efficient Fine-Tuning}
With the rapid scaling of large language models (LLMs), traditional full-parameter fine-tuning becomes progressively impractical due to exponentially increasing computational costs. Consequently, parameter-efficient fine-tuning (PEFT) techniques have assumed greater significance \cite{houlsby2019parameter}. There are two main paradigms for PEFT based on their principles: prompt-based tuning and adapter-based methods. 

Prompt-based techniques optimize the model through input-space interventions rather than architectural changes. Early implementations like Prompt Tuning \cite{lester2021powerscaleparameterefficientprompt} learn continuous task-specific embeddings prepended to input sequences, significantly reducing the computation of the parameters. Prefix-tuning \cite{li2021prefix} optimizes virtual token embeddings across all transformer layers, demonstrating improved capability on generation tasks. \citet{kwon2024stablepromptautomaticprompttuning} introduces adaptive proximal policy optimization to revise the stability and environment dependence. Although the Prompt-based techniques exhibit effective utilizations of few-shot and zero-shot data, they are still affected by their sensitivity to initialization and sequence length constraints.

Adapter-based approaches introduce small trainable modules between transformer layers, achieving parameter efficiency through freezing the base model. Preliminary methods \cite{houlsby2019parameter, mahabadi2021compacterefficientlowrankhypercomplex, bapna-firat-2019-simple, wang-etal-2021-k} introduce notable inference latency due to sequential computation bottlenecks. Low-Rank Adaptation (LoRA) further reduces the computational overhead through decomposing weight updates into low-rank matrices, achieving comparable performance to full fine-tuning with few parameters. Subsequent studies have extended this paradigm. AdaLoRA \cite{zhang2023adaloraadaptivebudgetallocation} dynamically allocates rank budgets across layers. LoRS \cite{hu2025lorsefficientlowrankadaptation} adopts weight recompute and computational graph rearrangement to reduce memory and computational consumption while improving performance. Moreover, \citet{gao2024higherlayersneedlora} observed that higher transformer layers require more LoRA experts to capture task-specific patterns, while lower layers exhibit significant redundancy. \citet{hu2024asloraadaptivesharinglowrank} enhance parameter efficiency by sharing the LoRA A matrix across all layers, further proof the significant parameter redundancy of conventional LoRA architecture.

\begin{figure*}[t]
    \centering
    \begin{subfigure}[b]{0.245\textwidth}
     \includegraphics[width=1\linewidth]{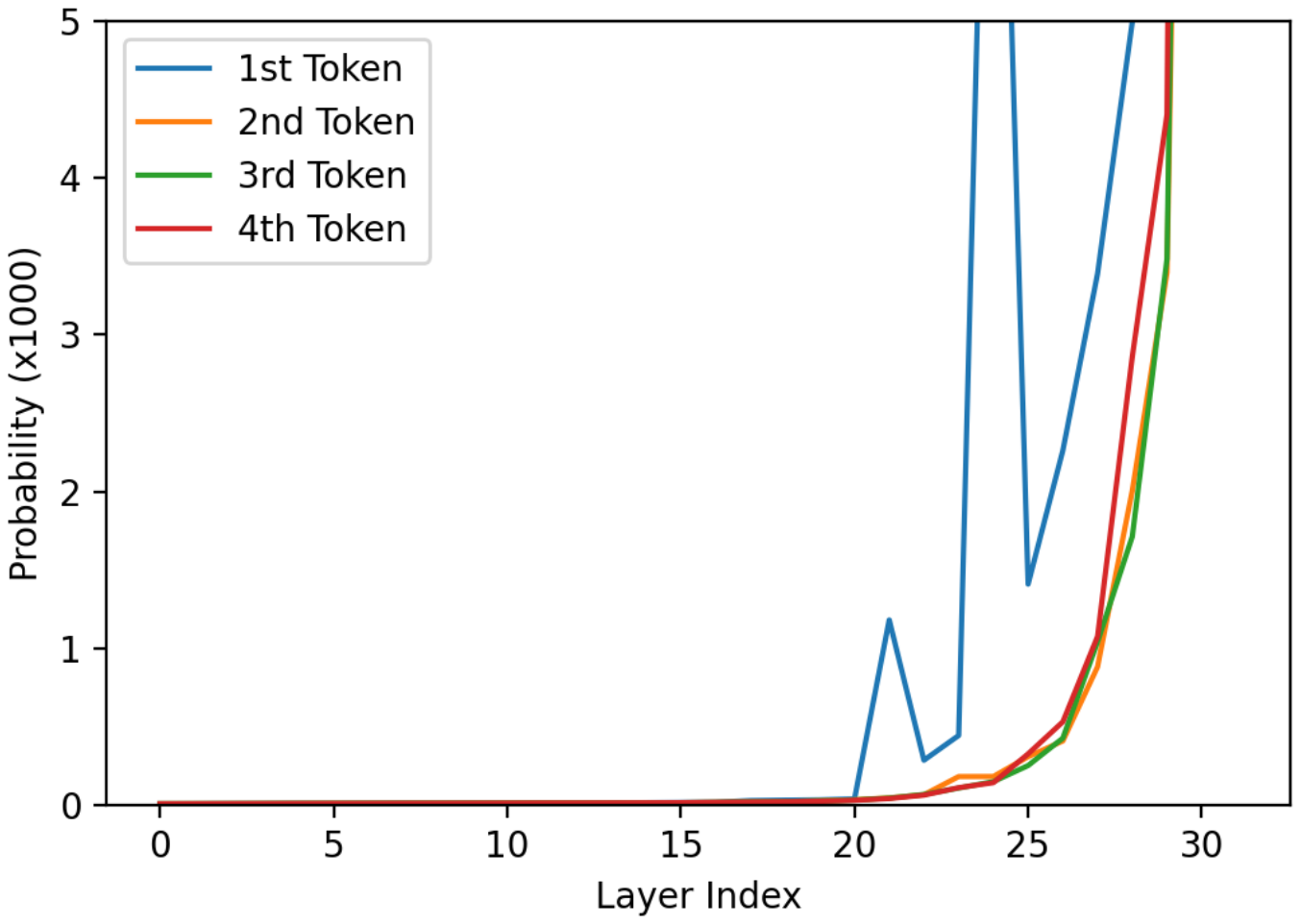}
    \caption{HotpotQA}
    \label{fig:sub1}
  \end{subfigure}
  \hfill 
  \begin{subfigure}[b]{0.245\textwidth}
     \includegraphics[width=1\linewidth]{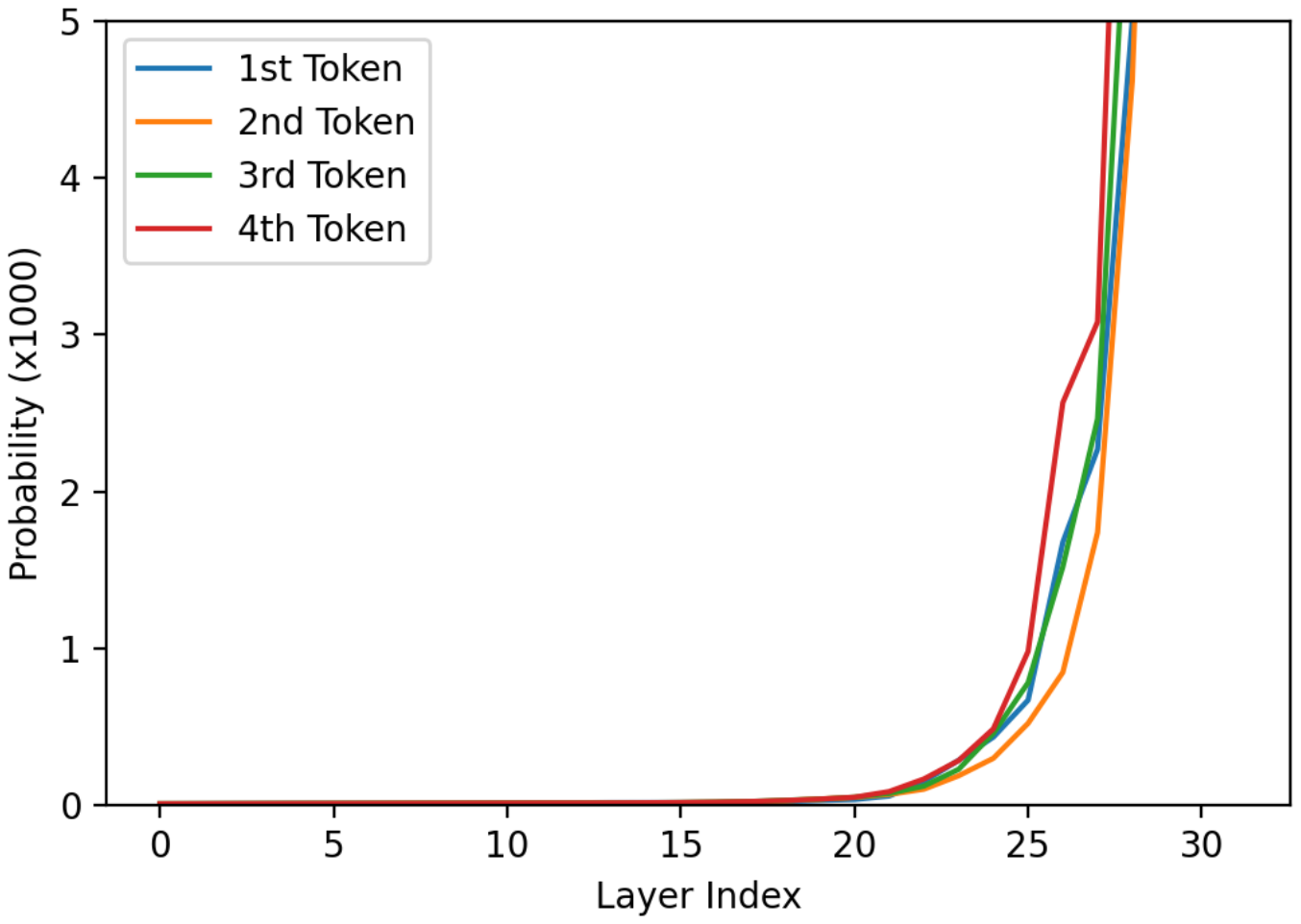}
    \caption{Samsum}
    \label{fig:sub2}
  \end{subfigure}
  \begin{subfigure}[b]{0.245\textwidth}
     \includegraphics[width=1\linewidth]{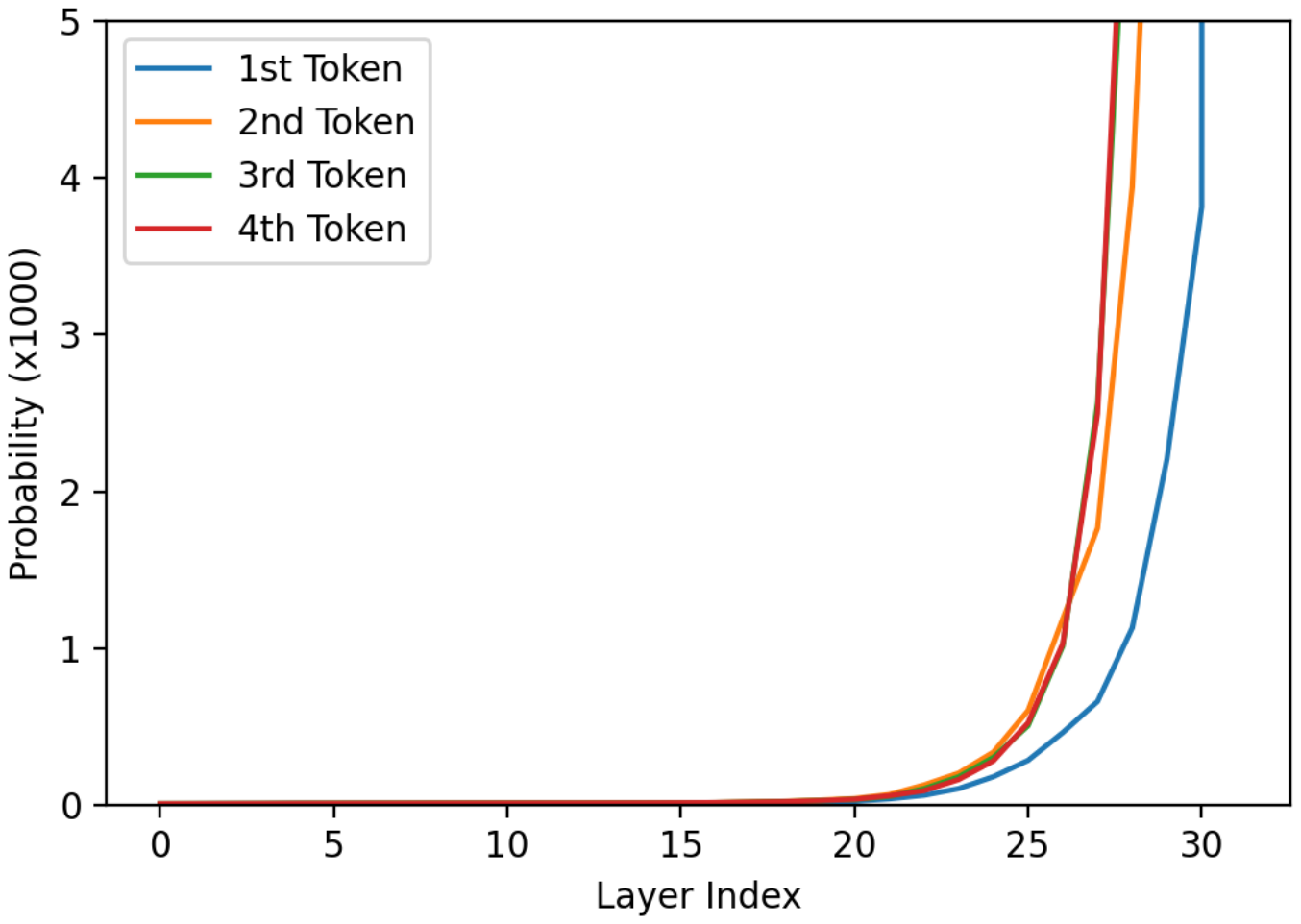}
    \caption{GSM8K}
    \label{fig:sub3}
  \end{subfigure}
  \begin{subfigure}[b]{0.245\textwidth}
     \includegraphics[width=1\linewidth]{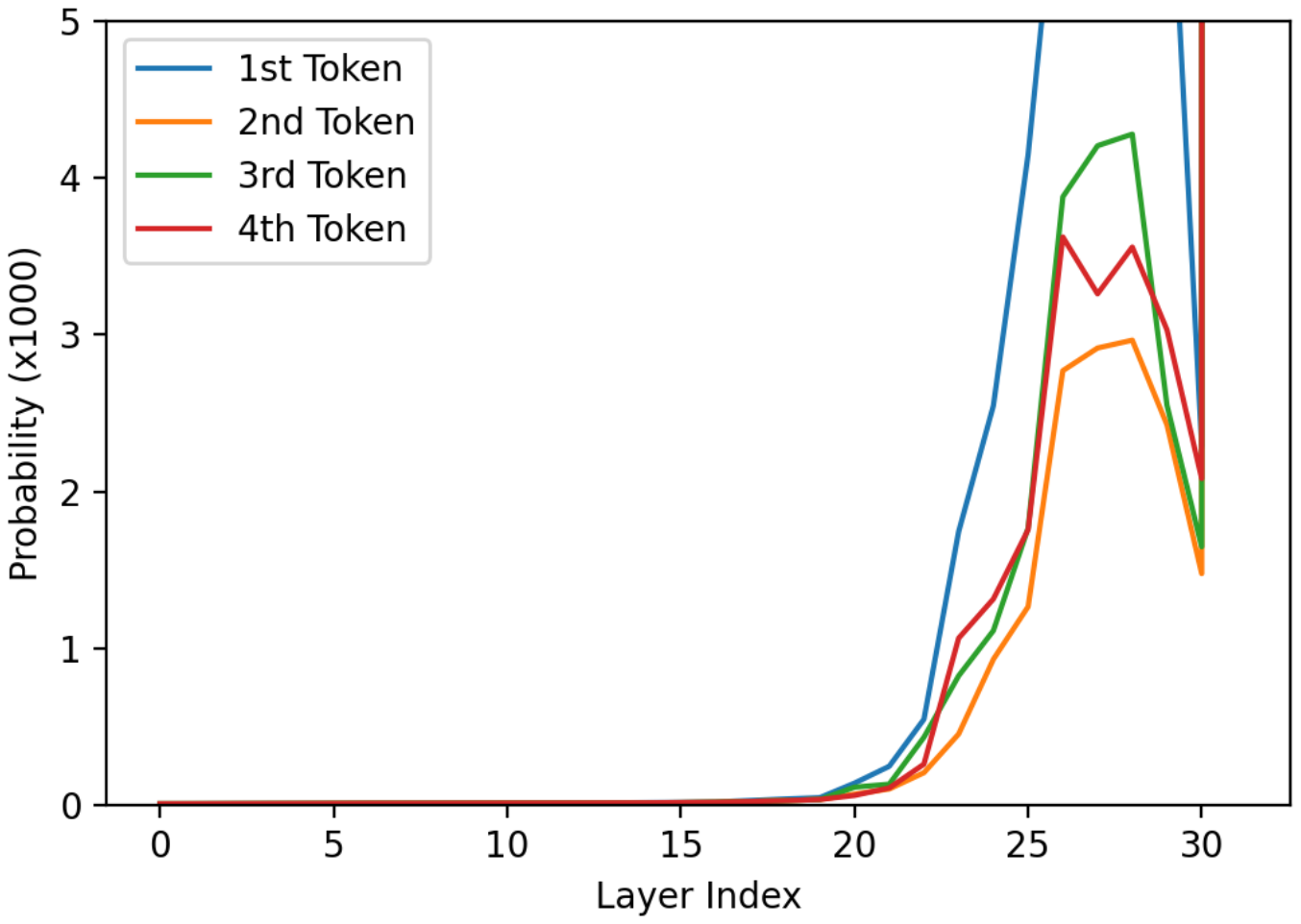}
    \caption{WMT23 (EN->ZH)}
    \label{fig:sub4}
  \end{subfigure}
    \caption{The average maximum probability of the first four tokens for each layer of Llama3.1-8B-Instruct model fine-tuned with LoRA on the four datasets.}
    \label{fig:1}
\end{figure*}

\begin{figure*}[t]
    \centering
    \begin{subfigure}[b]{0.245\textwidth}
     \includegraphics[width=1\linewidth]{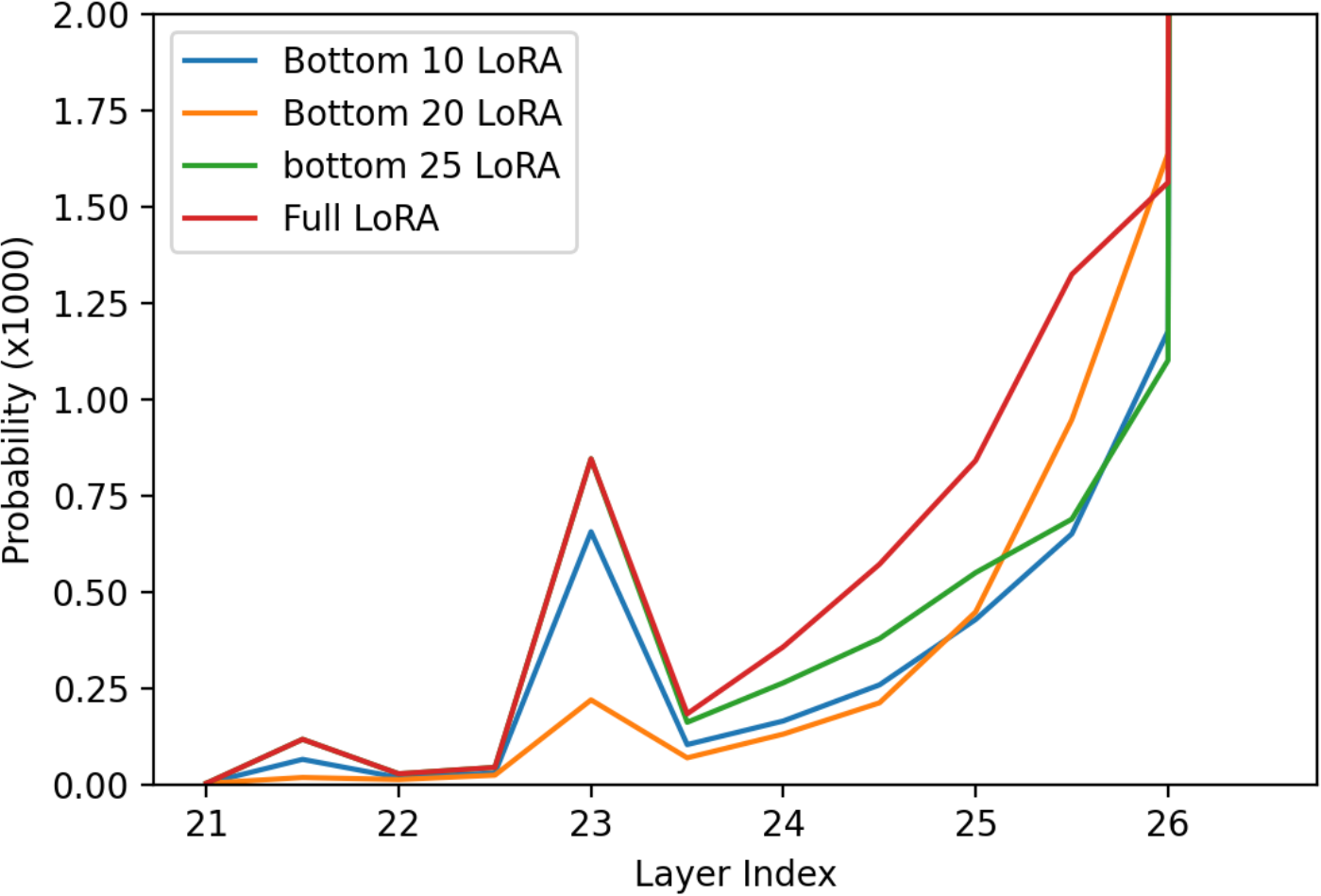}
    \caption{1st Token}
    \label{fig2:sub1}
  \end{subfigure}
  \hfill 
  \begin{subfigure}[b]{0.245\textwidth}
     \includegraphics[width=1\linewidth]{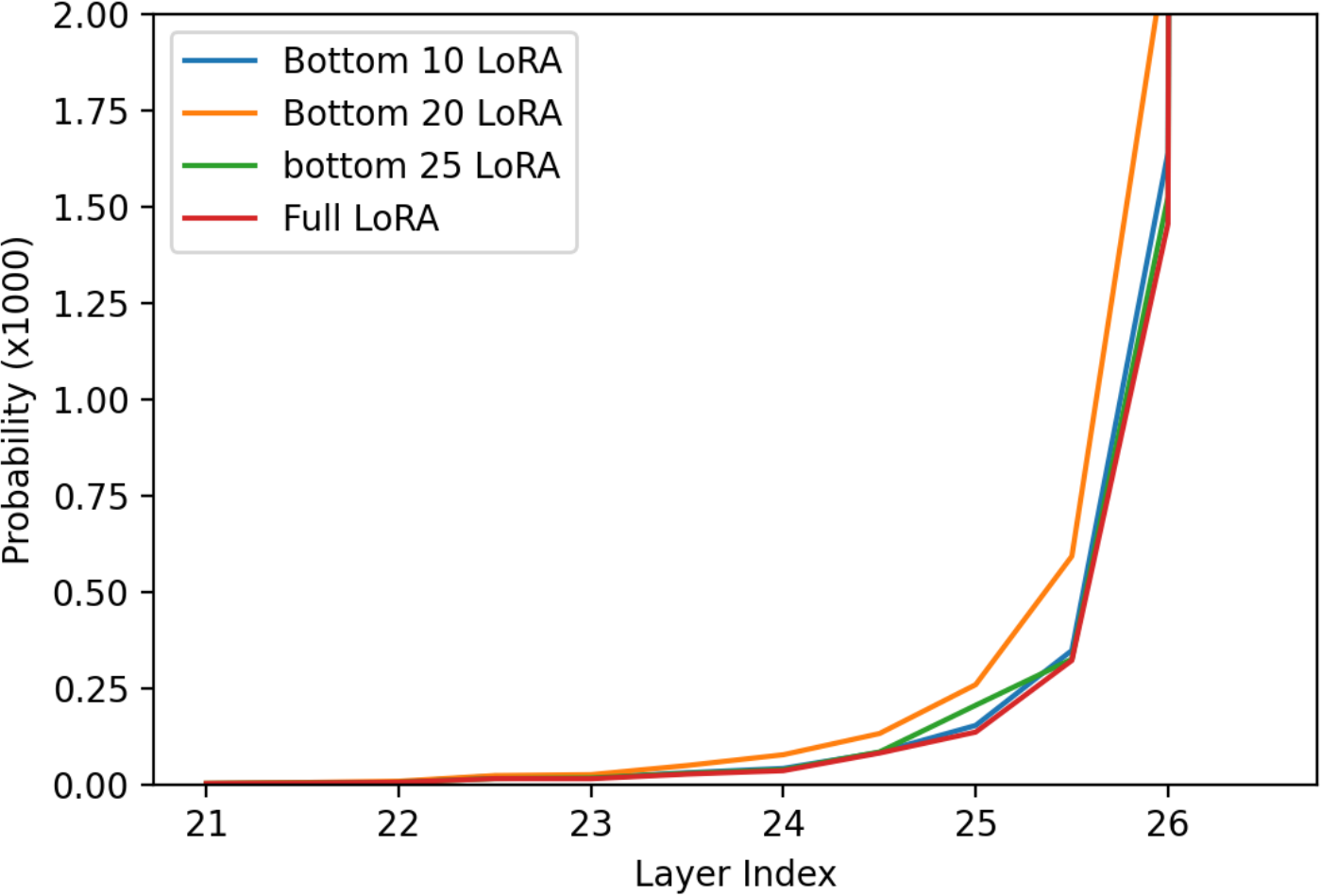}
    \caption{2nd Token}
    \label{fig2:sub2}
  \end{subfigure}
  \begin{subfigure}[b]{0.245\textwidth}
     \includegraphics[width=1\linewidth]{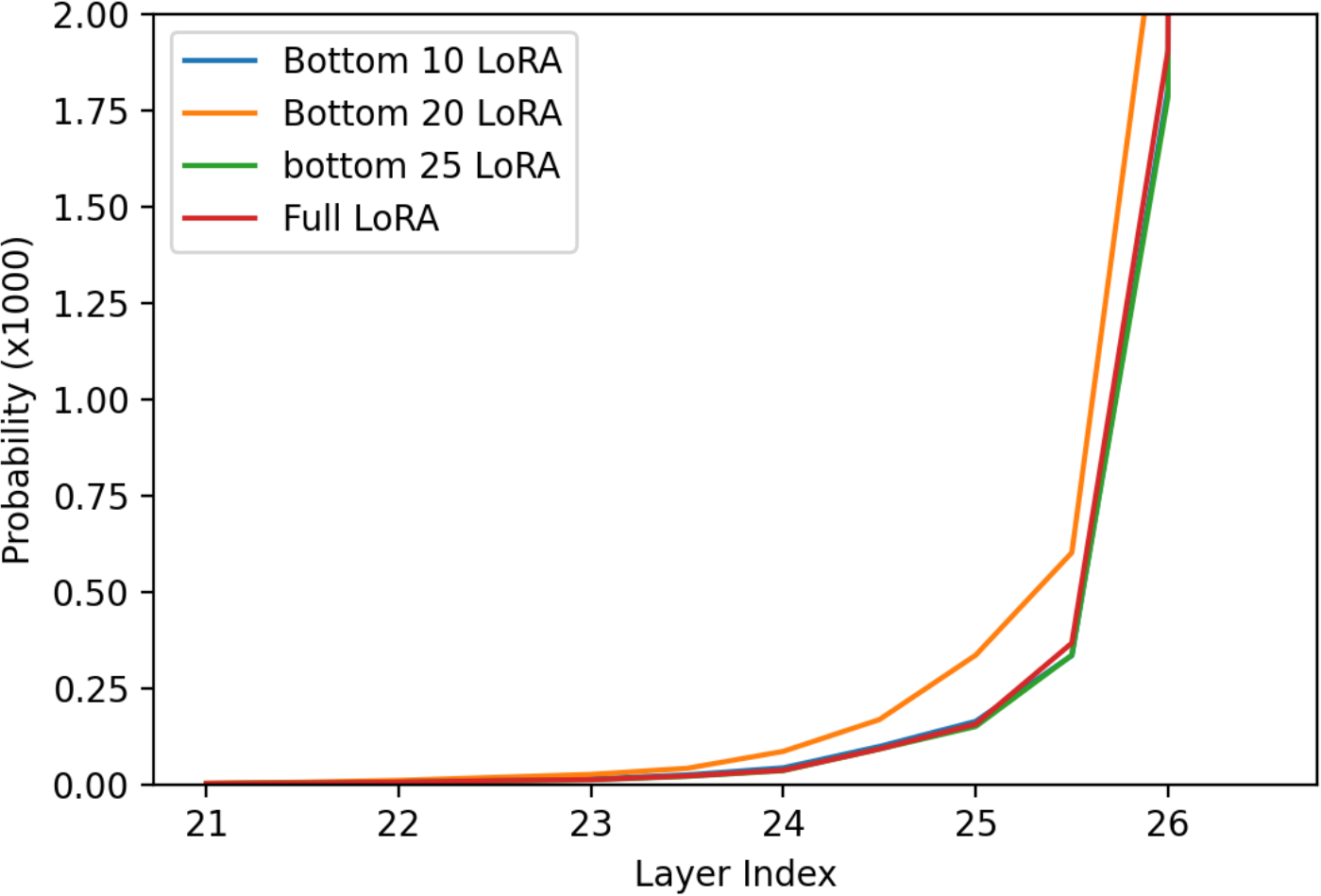}
    \caption{3rd Token}
    \label{fig2:sub3}
  \end{subfigure}
  \begin{subfigure}[b]{0.245\textwidth}
     \includegraphics[width=1\linewidth]{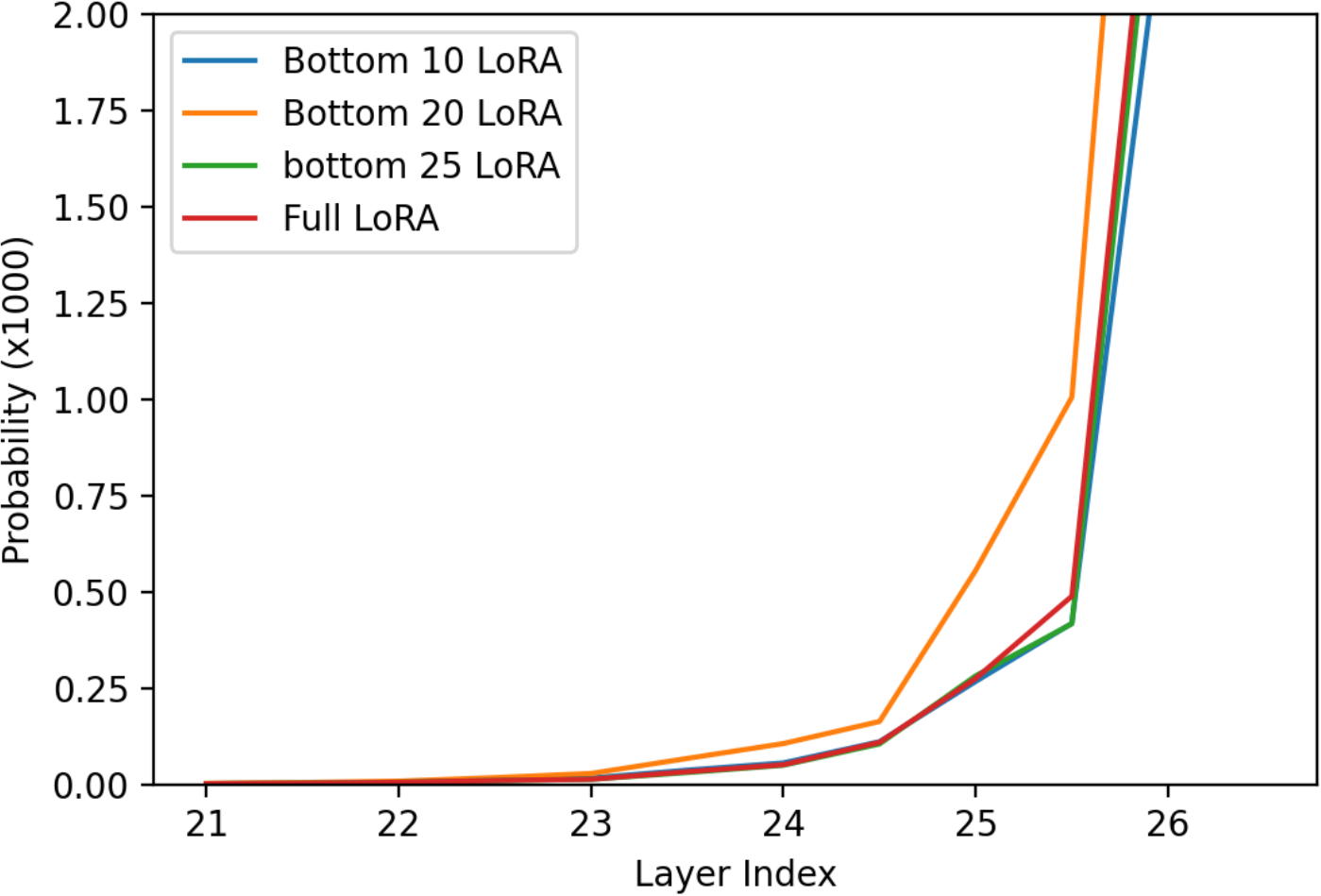}
    \caption{4th Token}
    \label{fig2:sub4}
  \end{subfigure}
    \caption{The average maximum probability of the first four tokens for each layer of Llama3.1-8B-Instruct model fine-tuned with LoRA on the HotpotQA dataset while dropping specific LoRA layers during inference.}
    \label{fig:2}
\end{figure*}

\subsection{Knowledge Distillation}
In contrast to the PEFT paradigm that focuses on optimizing model adaptation efficiency, an alternative trajectory explores knowledge distillation techniques for computational cost compression through inter-model knowledge transfer, which effectively reduces consumption, but generally faces the problem of a lack of task generalization abilities \cite{hahn2019self, sun2019patient, tang2019natural}. T\textit{f}-FD \cite{li2022self} enhances generalization through layer-wised self-distillation and student features reusing. However, the limited ability of the student model to represent complex tasks results in insufficient adaptability. \citet{liu2023normknowledgedistillationntoone} improves classification accuracy by linear transformation from $N$ to one, but neglects cross-layer semantic discrepancies and multilevel knowledge integration. However, our approach takes a fundamentally different way to further unleash the vintrinsic knowledge of LLMs, leveraging the information captured by the LoRA of the bottom layers while allowing the model to reason in its original, unmodified manner at the top layers without LoRA.

\section{Preliminary Analysis of LoRA} \label{sec.2}
In these preliminary analysis experiments, we use all three strong baselines fine-tuned with LoRA on four generation datasets in different tasks to explore the impact of LoRA for different layers.

\subsection{The Probability of Each Layer} \label{sec.2.1}

We randomly sampled 100 instances from the test set of our four datasets. Then we utilize fine-tuned LLMs to output the probability distribution across different layers. By encoding the ground truth for each sample and averaging the corresponding probabilities, we assessed whether the models arrived at the correct answers. For enhanced clarity in our analysis, we visualized the probability distributions of the first several tokens at each layer. Figure \ref{fig:1} and \ref{fig:4} illustrate these distributions for the Llama3.1-8B-Instruct and  Llama2-7B-Chat across the four datasets.

\paragraph{Observation} Our analysis reveals a distinct pattern in the probability distribution across the layers of LLMs. Specifically, the bottom layers exhibit relatively low and stable probabilities for first four tokens. However, at a certain ``boundary layer,'' we observe a sharp and significant increase in these probabilities. We posit that this transition reflects the model's progression from context comprehension and information extraction in the lower layers to answer formulation and refinement in the upper layers. This interpretation is supported by the visualization in Figure \ref{fig:4}, which demonstrates that the initial layers maintain consistently low probabilities, followed by a marked upward trend in the bottom layers. This abrupt shift suggests that the model begins to synthesize the extracted information and generate task-appropriate responses beyond this ``boundary layer''.

\begin{figure*}[t]
    \centering
     \includegraphics[width=1\linewidth]{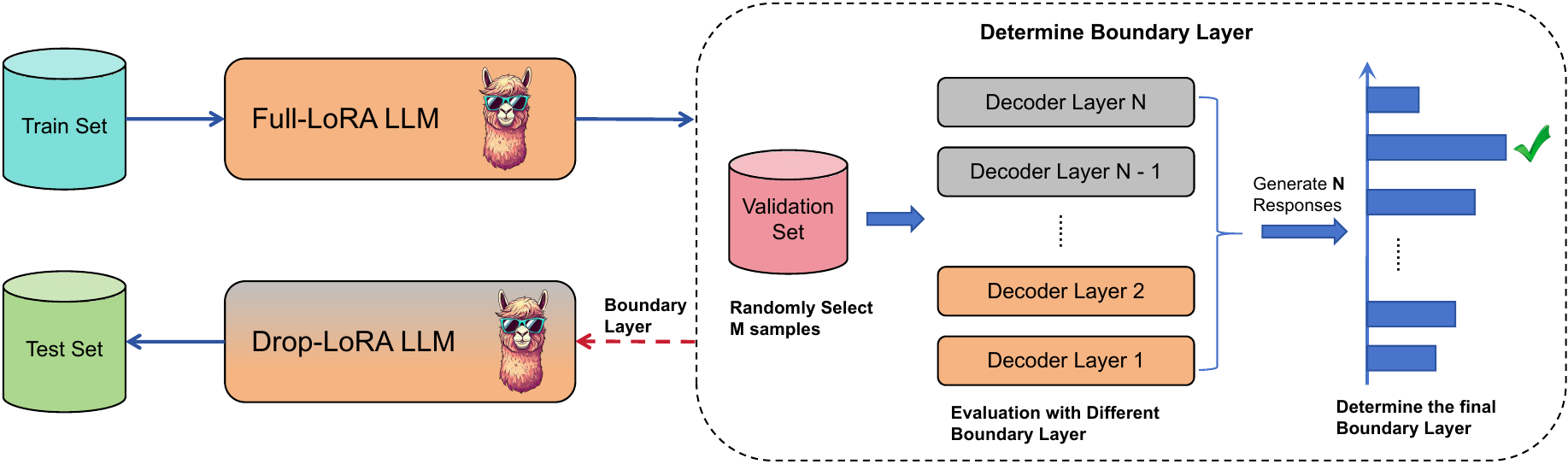}
    \caption{The overview of our proposed method.}
    \label{fig:6}
\end{figure*}

\subsection{Top LoRA Are Not Necessary}
Building on the conclusions drawn in Section \ref{sec.2.1}, we posit that for downstream tasks, it is crucial for LLMs to understand the context and effectively capture information. Imposing rigid formatting constraints may negatively impact model performance, given that LLMs are primarily pre-trained for text completion on natural language corpora. To investigate the layers of LoRA that are essential, we conduct experiments by selectively dropping specific LoRA of layers from an LLM. Specifically, we randomly select 100 samples from the test set and visualize the average maximum probability of the first four tokens at each layer when keeping the LoRA of the bottom 10, 20, and 25 layers. The resulting curves are depicted in Figure \ref{fig:2}. We plot the average maximum probability of the same token across various scenarios in each sub-figure.

\paragraph{Observation} Dropping the LoRA of the specific numbers of top layers does not significantly affect the output probabilities. As illustrated in Figure \ref{fig:2}, keeping the LoRA of bottom layers does not significantly affect the model's output probabilities too much, particularly for the 2nd to 4th tokens (Figure \ref{fig2:sub2}, \ref{fig2:sub3}, \ref{fig2:sub4}). This observation is reasonable, as the first token typically determines the response pattern for downstream tasks, making it more susceptible to fluctuations compared to subsequent tokens. This aligns with the findings of \citet{zhan-etal-2024-prefix}, which suggests that fine-tuning primarily impacts the first token. Furthermore, we observe that when keeping the LoRA of the bottom 20 layers (depicted by the orange curve), the model's maximum output probabilities outperform those in other configurations, including the baseline. Notably, even for the first token's probability, this ``boundary layer'' also results in higher maximum output probabilities in the last several layers compared to the baseline.

\section{Method}
Building upon our hypothesis and the empirical observations detailed in Section \ref{sec.2}, we propose a simple yet effective strategy to improve model performance without requiring additional fine-tuning. Our approach centers on the removal of LoRA components from layers situated above an identified ``boundary layer.'' A comprehensive overview of this methodology is illustrated in Figure \ref{fig:6}.

Initially, we apply LoRA to all layers to conduct supervised fine-tuning (SFT) of a large language model (LLM) on a downstream task, adhering to standard procedures. Subsequently, we identify the ``boundary layer'' where LoRA should be removed. As analyzed in Section \ref{sec.2}, a computationally efficient method to determine this layer involves calculating the average probability of the ground truth from the logits output of each layer and visualizing these probabilities. The point where the curve begins to rise is manually identified as the ``boundary layer.'' After removing the LoRA from layers beyond this point, we can directly generate responses for the test set without further training.

Additionally, we propose a more precise and automated method to identify the ``boundary layer,'' as illustrated in Figure \ref{fig:6}. After fine-tuning the LLMs with full LoRA, we randomly select a set of $M$ samples from the validation set to evaluate model performance after dropping LoRA at different ``boundary layers.'' For instance, in the case of the \textbf{Llama3.1-8B-Instruct} model, which has 32 decoder layers, we obtain 32 evaluation results corresponding to different ``boundary layers.'' We then select the best-performing result as our final ``boundary layer'' and evaluate the model performance on the test set after removing LoRA beyond this point. By using validation set samples to determine the boundary layer and subsequently evaluating on the test set, we mitigate the risk of overfitting the boundary layer to a specific dataset, thereby enhancing the credibility of our results.

\section{Experiments}

\begin{table*}[t]
  \centering
    \renewcommand\arraystretch{1}
    \tabcolsep=0.12cm
    \scalebox{1}{
    \begin{tabular}{lcccccc}
    \toprule
     & \multicolumn{2}{c}{\textbf{HotpotQA}} & \textbf{GSM8K} & \multicolumn{2}{c}{\textbf{Samsum}} & \multicolumn{1}{c}{\textbf{WMT23(En->Zh)}} \\
     \midrule
     Model& EM  & F1 & EM  & ROUGE & GPT-Score & BLEU \\
      \midrule
      Phi-2 (Fine-tuned) & 62.8 & \textbf{70.0}  & 57.1 & \textbf{33.7} & 71.2 &   \textbf{6.4} \\
      Phi-2 (Ours) & \textbf{65.6} & 68.4 & \textbf{57.8} & 32.5 & \textbf{72.5} &  5.4 \\
      \midrule
      Llama2-7B-Chat (Fine-tuned) & 66.2 & \textbf{73.5} &  36.1 &   42.8 & 78.0 & 27.0  \\
      Llama2-7B-Chat (Ours) & \textbf{69.1} & 71.6 & \textbf{38.6} & \textbf{43.5} & \textbf{80.3} & \textbf{27.4}  \\
      \midrule
      Llama3.1-8B (Fine-tuned) & 73.1 & \textbf{80.4}   &  73.1 & \textbf{38.2}  & 79.5 &  33.0   \\
      Llama3.1-8B (Partial Fine-tuned) & 73.0 & 79.9 & 73.9 & 37.7 & 79.8  &  33.3 \\
      Llama3.1-8B (Ours) & \textbf{74.1} & 80.3 & \textbf{74.3} & 38.1 &  \textbf{80.6} & \textbf{35.8}  \\
      
      \bottomrule
    \end{tabular}}
    \caption{The results (\%) on the test set of the five datasets in our experiments. \textbf{Bold} numbers indicate the better result for each baseline.}
  \label{table1}
\end{table*}

\subsection{Setup} \label{sec.4.1}
\paragraph{Dataset} We use four widely used datasets in different generation tasks in our experiments: \textbf{HotpotQA} \cite{yang-etal-2018-hotpotqa} for multi-document question answering, \textbf{GSM8K} \cite{cobbe2021training} for mathematical reasoning, \textbf{Samsum} \cite{gliwa-etal-2019-samsum} for text summarization, and \textbf{WMT23} \cite{kocmi-etal-2023-findings} for machine translation. The statistic about them is shown in Table \ref{tab:1}. 

\paragraph{Evaluation Metric}
We employ different evaluation metrics for these four tasks seperately. We employ Exact Match (EM) and F1 score to evaluate the performance of the HotpotQA dataset, whose purpose is to measure how completely the generated response contains the label. For \textbf{GSM8K}, we only use the EM score to evaluate whether the final calculated results is correct. For \textbf{Samsum} and \textbf{WMT23}, we use the traditional ROUGE-L and BLEU\footnote{https://github.com/mjpost/sacrebleu} \cite{post-2018-call} seperately at first. However, these two tasks can have similar meanings in different expressions. Thus, we also design a GPT-Score (Appendix \ref{app.2}) to evaluate the generation output from various perspectives, such as fluency, accuracy, or readability, by asking the LLMs to act as a human. 

\begin{table}[t]
    \centering
    \renewcommand\arraystretch{0.85}
    \scalebox{1}{
    \begin{tabular}{cccc}
        \toprule
        Dataset & Train & Validation & Test \\
        \midrule
        HotpotQA & 50,000 & 7,405 & 7,405  \\
        GSM8K & 7,473 & 1,319 & 1,319  \\
        Samsum & 14,732 & 818 & 819 \\
        WMT23 & 50,000 & 500 & 2,074 \\
        \bottomrule
    \end{tabular}}
    \caption{The statistics of the four datasets we used in our experimental setting.}
    \label{tab:1}
\end{table}

\paragraph{Model} We evaluate our method on three strong baselines: \textbf{Phi-2}\footnote{https://huggingface.co/microsoft/phi-2}  \cite{textbooks2}, \textbf{Llama2-7B-Chat}\footnote{https://huggingface.co/meta-llama/Llama-2-7b-chat-hf} \cite{touvron2023llama}, and \textbf{Llama-3.1-8B-Instruct}\footnote{https://huggingface.co/meta-llama/Llama-3.1-8B-Instruct} \cite{dubey2024llama}. More details of implementation will be shown in Appendix \ref{app.3}.

\subsection{Main Results} \label{sec.5.2}
As shown in Table \ref{table1}, our method of selectively dropping LoRA after specific layers outperforms the baseline in nearly all experiments. For the HotpotQA dataset, our method achieves a higher EM score. However, it consistently underperforms all three baselines in terms of F1 scores. This suggests that our approach is more effective in generating responses that contain the correct answers compared to the baselines. As discussed in our previous analysis, the LoRA components in the lower layers primarily help the LLM learn to format answers according to specific downstream tasks. When these LoRA components are dropped, the model continues to reason like the original model, often generating answers in natural language. This behavior leads to responses that include additional words alongside the correct answer, thereby negatively impacting the F1 scores.

For the GSM8K and WMT23 datasets, our method mostly outperforms the baselines, further verifying that removing LoRA from several top layers can enhance the reasoning capability of LLMs. It should be noted that the performance of our method on the Samsum dataset drops slightly on two baselines while achieving a better GPT-Score. This is because our approach deals with summarization tasks that are highly open-ended, and our method reduces the constraints imposed by the LoRA of top layers on the LLMs' inference capabilities, allowing the generated summary to be more comprehensive. However, as a trade-off, it also loses the ability of the top LoRA to mimic training data for formatting the answers, leading to a decrease in Rouge scores. Regarding the decline in performance on the WMT23 test set with Phi-2, our analysis revealed that the training data remains too challenging for Phi-2. We believe this is why removing LoRA in Phi-2 can not get the same improvement as the other two models.

\begin{figure}[t]
    \centering
     \includegraphics[width=1\linewidth]{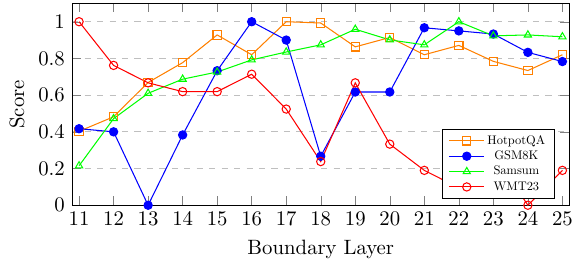}
    \caption{The performance of different ``boundary layer'' of Llama3.1-8B-Instruct model. The \textbf{Score} means the corresponding automatic evaluation metric of four datasets.}
    \label{fig:3}
\end{figure}

\subsection{Ablation Study}
In this work, we present a robust and unified approach that resists decomposition into discrete components for analysis. However, our strategy of selectively dropping LoRA beyond the ``boundary layer'' during inference introduces a challenge: the need for a two-step process involving fine-tuning followed by LoRA removal. We conducted experiments across four datasets using Llama3.1-8B-Instruct, applying LoRA only to the bottom 20 layers of the LLMs during the fine-tuning phase. The evaluation results are shown in Table \ref{table1}, which is the row of "Partial Fine-tuned". Interestingly, the performance of LLMs fine-tuned with LoRA in the bottom 20 layers closely matches that of the baseline models. This suggests that, regardless of whether LoRA is applied to all layers during fine-tuning, some LoRA components are particularly effective at capturing and understanding context, while others excel at synthesizing and refining answers to suit downstream tasks. These findings highlight the necessity of our two-step process.

\section{Analysis}

\subsection{Different Boundary Layer}
To provide a more intuitive rationale for our selection of the ``boundary layer'', we conduct additional experiments by removing the LoRA with different ``boundary layers''. Specifically, we assessed model performance after dropping the LoRA after the K=$[10, 25]$ layers, with the results illustrated in Figure \ref{fig:3}. Consistently, removing the LoRA after the 15-20 layer resulted in better performance across all four datasets. This finding highlights the robustness of our approach and suggests that the ``boundary layer'' for one model is likely to remain a range across different downstream tasks.
\begin{figure}[t]
    \centering
     \includegraphics[width=1\linewidth]{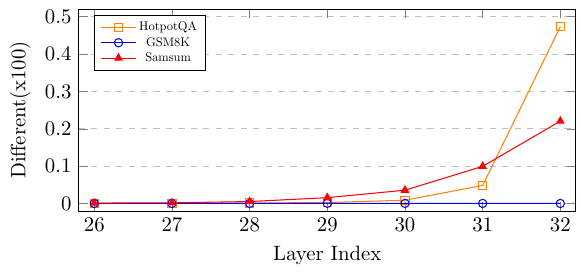}
    \caption{The probability difference of ground truth between our method and baseline on three datasets.}
    \label{fig:7}
\end{figure}

\subsection{Post-Hoc Analysis}
Furthermore, we conduct some experiments to investigate how dropping LoRA affecting the LLMs performance and verify its effect. Specifically, we evaluated the probability distribution of the first token generated by our method and baseline models across three datasets: HotpotQA, Samsum, and GSM8K. Specifically, we extracted and compared the average probability of the ground truth token for all decoding layers. Typically, the average probability of the ground truth token for the first token is relatively low, making visualization challenging. To address this, we calculated the difference in average probabilities for each layer between our method and the baseline, and visualized the results in Figure \ref{fig:7}. The visualization reveals that after dropping certain LoRA components, our method tends to assign higher probabilities to the ground truth token, especially near the output layers. It is important to note that the probability differences for GSM8K remain stable. This stability arises because, in GSM8K, the model must perform reasoning before arriving at the final answer, resulting in low probabilities for the first token in both our method and the baseline. This contrasts with HotpotQA, where the ground truth is relatively shorter, leading to the largest probability differences. A case study will be introduced in Appendix \ref{app.1} to concretely show the effect of our method.

\subsection{Comparison of Labeling and Generation}
Besides evaluating the performance of generation tasks, we also conduct analysis to verify whether our method can improve the performance on labeling tasks, where LLMs also get significant performance. The HotpotQA dataset can be partially viewed as a labeling task, featuring two types of questions: Bridge and Comparison. The answers to comparison questions are typically ``Yes'', ``No'', or an choice between ``A'' or ``B'', which is similar to the binary classification. We begin by assessing the performance improvements on these tasks. As depicted in Table \ref{table6}, our approach yields EM scores for both question types, albeit with a reduction in F1 scores, mirroring trends observed in our main results (Section \ref{sec.5.2}). In particular, our method shows approximately double the improvement in EM scores for bridge compared to comparison.

Besides HotpotQA, we further evaluate our method using a purely labeling dataset. We fine-tune the LLaMA3.1-8B-Instruct with LoRA on a subset of the \textbf{E-Commerce} dataset \cite{zhang2018dua}, selecting 50,000/500/2,000 data for the train/validation/test set, equally divided between positive and negative samples. This dataset focuses on dialogue response selection \cite{CHEN2024111687}, requiring models to ascertain the suitability of a given response within a dialogue context. We utilize accuracy as the performance metric, pre-processing the data to align with LLM natural language formats with "Yes" or "No" labels for fine-tuning and evaluation. The results, presented in Table \ref{table7}, indicate that our method surpasses the baseline by 1.2\%, suggesting its efficacy for classification tasks. In addition, the output for this data should always be a single word. However, we observed that 13\% of samples exceeded one word after dropping LoRA, with the additional content providing explanatory reasoning for the label. This observation supports our hypothesis that certain top LoRA are primarily involved in refining and formatting responses.

\begin{table}[t]
  \centering
    \renewcommand\arraystretch{1}
    \tabcolsep=0.5cm
    \scalebox{1}{
    \begin{tabular}{lcc}
    \toprule
     Question Type & EM  & F1  \\
     \midrule
    Bridge & 78.1 & \textbf{81.7} \\
      Bridge (Ours) & \textbf{79.7}  & 80.4  \\
      \midrule
      Comparison & 71.9 & \textbf{79.8}  \\
      Comparison (Ours) & \textbf{72.7} & 78.7  \\
      \bottomrule
    \end{tabular}}
    \caption{The results of different types of question (\%) HotpotQA dataset with Llama3.1-8B-Instruct fine-tuned with LoRA. \textbf{Bold} numbers indicate the best results.}
  \label{table6}
\end{table}

\begin{table}[t]
  \centering
    \renewcommand\arraystretch{1}
    \tabcolsep=0.32cm
    \scalebox{1}{
    \begin{tabular}{lc}
    \toprule
    Method &  Accuracy (\%)  \\
     \midrule
    E-Commerce (Fine-tuned) & 68.6  \\
      E-Commerce   (Ours) &  \textbf{69.8} \\
      \bottomrule
    \end{tabular}}
    \caption{The accuracy (\%) of our method and baseline on Llama3.1-8B-Instruct fine-tuned with E-Commerce dataset. \textbf{Bold} numbers indicate the best results.}
  \label{table7}
\end{table}

\begin{table}[t]
    \centering
    \renewcommand\arraystretch{1}
    \tabcolsep=0.32cm
    \scalebox{1}{
    \begin{tabular}{lcc}
        \toprule
         & Samsum & WMT23(EN->ZH) \\
        \midrule
         \multicolumn{3}{c}{LLMs-based Evaluation} \\
        \midrule
        Baseline & 37.6  & 44.2   \\
        Ours & \textbf{62.4} & \textbf{55.8}  \\
        \midrule
         \multicolumn{3}{c}{Real Human Evaluation}  \\
        \midrule
        Baseline &  10.0 &  43.3  \\
        Ours & \textbf{90.0} &  \textbf{56.7} \\
        \bottomrule
    \end{tabular}}
    \caption{The LLMs-based and real human evaluation on Llama3.1-8B-Instruct. This percentage(\%) refers to the proportion of all test data where the output of this method is better than that of another method. \textbf{Bold} numbers indicate the best results.}
    \label{table3}
\end{table}

\begin{table*}[t]
  \centering
    \renewcommand\arraystretch{1}
    \tabcolsep=0.32cm
    \scalebox{1}{
    \begin{tabular}{lcccccc}
    \toprule
     & \multicolumn{2}{c}{\textbf{HotpotQA}} & \textbf{GSM8K} & \multicolumn{2}{c}{\textbf{Samsum}} & \multicolumn{1}{c}{\textbf{WMT23(En->Zh)}} \\
    \midrule
     Method & EM  & F1 & EM & ROUGE & GPT-Score & BLEU \\
     \midrule
    Baseline ($r=16$) & 73.9 & \textbf{80.1} & 74.4 & \textbf{38.6} & 80.4 & 32.9   \\
      Ours ($r=16$)  & \textbf{74.6} & \textbf{80.1} & \textbf{76.4} & 38.0 & \textbf{80.7} & \textbf{36.8}   \\
      \midrule
      Baseline ($r=32$) & 72.9 & 79.9 & 75.8 & 38.5 & 80.8 & 32.6  \\
      Ours ($r=32$)  & \textbf{73.6} & \textbf{80.0}  & \textbf{76.0} & \textbf{38.6} & \textbf{81.2} &  \textbf{36.0} \\
      \bottomrule
    \end{tabular}}
    \caption{The results of different LoRA rank on Llama3.1-8B-Instruct. \textbf{Bold} numbers indicate the best results.}
  \label{table11}
\end{table*}

\subsection{Generation Quality} \label{sec6.4}
Generally speaking, directly modifying some parameters of a fine-tuned LLM without continuing training may result in the generated content being incoherent or garbled. Thus, we further design two strategies to evaluate the generation quality of our method on \textbf{Samsum} and \textbf{WMT23}, since these two tasks are open-ended generation tasks and place greater emphasis on the quality of the generated results. The first one is similar to GPT-Score, which we ask LLMs to act as humans to choose the better one between our method and baselines. The other is real human evaluation. We randomly sample 30 examples seperately in test set of \textbf{Samsum} and \textbf{WMT23} datasets, and recruited three computer science students proficient in both English and Chinese to choose the better approach between our method and the baseline method based on four aspects: fluency, accuracy, readability, and completeness. Then, we calculated the average results from these three students. As shown in Table \ref{table3}, the percentage(\%) represents the proportion of all test data where the evaluators considered the results of this method to be better. 

The output from our method gets a higher proportion consistently, whether evaluated by LLMs or real humans. It indicates that our method can improve the model performance without affecting the generation quality through only dropping some LoRA during inference. The evaluation instruction for LLMs will be shown in Table \ref{table9}. In comparison to text summarization, our method did not achieve as significant results in translation tasks. We think it is because the prediction results in translation tasks are more strongly constrained by the source sentence, which means that the gains from dropping LoRA of specific layers are relatively smaller. This also indicates that our method performs better in tasks with fewer constraints on the ground truth format, such as open-ended generation tasks.

\subsection{Impact of LoRA Rank} \label{sec 6.5}
\guanhua{One of the pivotal factors influencing the performance of LoRA tuning is the choice of LoRA rank. As different rank values can lead to varying performances even with identical models and datasets, we conducted a comprehensive experimental evaluation on the Llama3.1-8B-Instruct model. Specifically, we examined rank values of $\{16,32\}$, with the detailed results presented in Table \ref{table11}. Our empirical findings demonstrate that our method consistently outperforms nearly all baseline approaches, thereby providing strong evidence for its generalization capabilities. Furthermore, our analysis revealed that while overfitting phenomena were observed on the GSM8K and WMT23 datasets, as indicated by performance degradation, dropping top LoRA can effectively mitigate this issue and enhance model performance. It means that our method possesses inherent mechanisms that can alleviate overfitting to a significant extent, making it more robust across diverse datasets and tasks.}

\subsection{Robustness of OOD Data}
Building on the insights from Section \ref{sec 6.5}, we further investigate our method's robustness through out-of-domain (OOD) evaluation. To this end, we construct two test sets by randomly sampling 2,000 en→zh parallel sentences from \textbf{WikiMatrix} \cite{schwenk-etal-2021-wikimatrix} (general domain) and \textbf{ParaMed} \cite{liu2021paramed} (biomedical domain). These datasets are evaluated on Llama3.1-8B-Instruct models fine-tuned exclusively on WMT23 (news domain), using the same boundary layer configuration. As shown in Table \ref{table12}, our method demonstrates consistent superiority over baselines despite potential domain mismatch, suggesting its strong generalization capabilities even when the optimal boundary layer may vary across domains.

\begin{table}[t]
  \centering
    \renewcommand\arraystretch{1}
    \tabcolsep=0.65cm
    \scalebox{1}{
    \begin{tabular}{lcc}
    \toprule
     & {\textbf{WIKI}} & {\textbf{PARAM}} \\
    \midrule
    Baseline & 19.9 &  22.6  \\
      Ours  & \textbf{21.0} & \textbf{24.9}  \\
      \bottomrule
    \end{tabular}}
    \caption{The BLEU score of different domain translation data on Llama3.1-8B-Instruct fine-tuned with News data. \textbf{Bold} numbers indicate the best results.}
  \label{table12}
\end{table}
\subsection{Error Analysis}
In this section, we examine cases where our proposed method underperforms across all experimental setups. Our results in Table \ref{table1} indicate a consistent decline in F1 scores compared to the baseline. Thus, we conducted a detailed error analysis on the HotpotQA dataset, focusing on samples where our method yielded lower F1 scores. Our analysis reveals two key observations: (1) In 56\% of these cases, the output generated by our method fully contains the baseline’s output, and (2) when both methods produce correct answers, this overlap increases to 96\%. This suggests that our approach tends to generate longer and more comprehensive responses compared to the baseline. We hypothesize that this behavior stems from the influence of top LoRA in LLMs. These LoRA appear to bias the model toward producing more structured or formatted outputs. When these layers are removed during inference, the output probability distribution becomes smoother, potentially leading to more verbose and inclusive responses.

\section{Conclusion}
In this work, we investigate the impact of fine-tuning LoRA across various layers in LLMs, revealing that LoRA beyond a specific "boundary layer" becomes redundant during inference. We propose a simple yet effective method to enhance LoRA-fine-tuned LLM performance. Future directions include leveraging these findings to develop advanced techniques for optimizing LoRA efficacy.

\section{Limitation}
Our approach requires sampling a subset from the validation set to identify the boundary layer, introducing additional computational cost, although the boundary layer can be directly set to $k=\{15,20\}$ as a practical alternative based on our analysis. Our experimental results highlight that the method imposes specific requirements on the capability of LLMs: the model must possess sufficient capacity to effectively learn the downstream task. Otherwise, the LoRA of all layers may be utilized during inference to compensate for insufficient task-specific knowledge. For example, a model like Phi-2, which achieves only a 6.4 BLEU score on translation tasks due to inadequate learning, would significantly undermine the efficacy of our approach.


\bibliography{acl_latex}

\appendix

\section{Appendix}
\label{sec:appendix}
\subsection{Case Study} \label{app.1}
We analyze several random samples concretely to explore the effect of our proposed method. The most notable results are observed with the \textbf{Samsum} dataset, because the summarization task is very flexible, where multiple valid expressions can effectively capture the essence of a paragraph. The examples provided in Table \ref{table10}, after applying our method to drop specific LoRA, the LLM generates responses that are more comprehensive and accurate. This observation aligns with our initial hypothesis: the LoRA of the top several layers primarily functions to format responses according to downstream task, which may, to some extent, constrain the model's reasoning capabilities. By removing these specific LoRA, the LLM is able to leverage the knowledge captured with the bottom LoRA to engage in more divergent reasoning.

\begin{figure*}[t]
    \centering
  \begin{subfigure}[b]{0.245\textwidth}
     \includegraphics[width=1\linewidth]{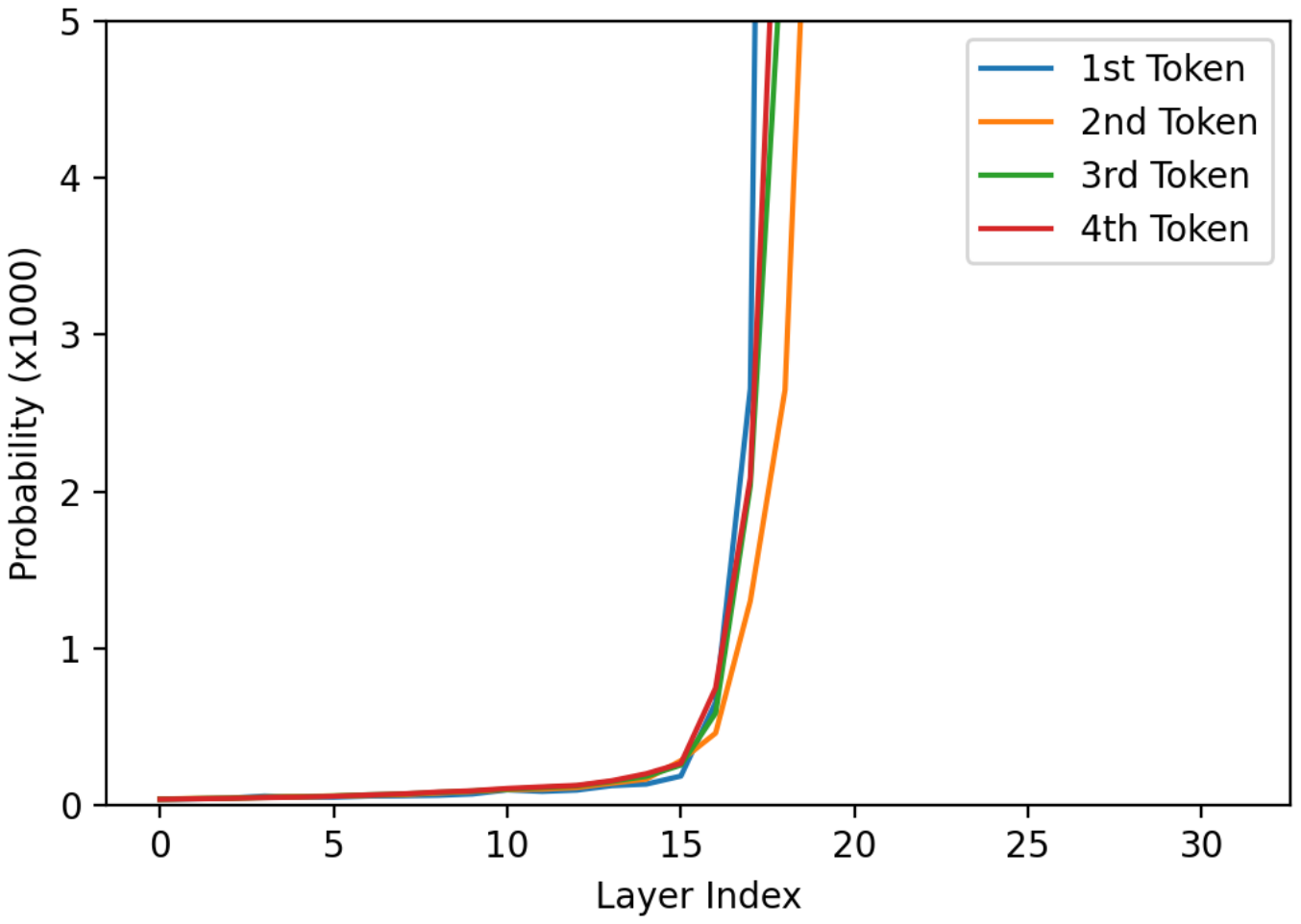}
    \caption{HotpotQA}
    \label{fig4:sub1}
  \end{subfigure}
  \begin{subfigure}[b]{0.245\textwidth}
     \includegraphics[width=1\linewidth]{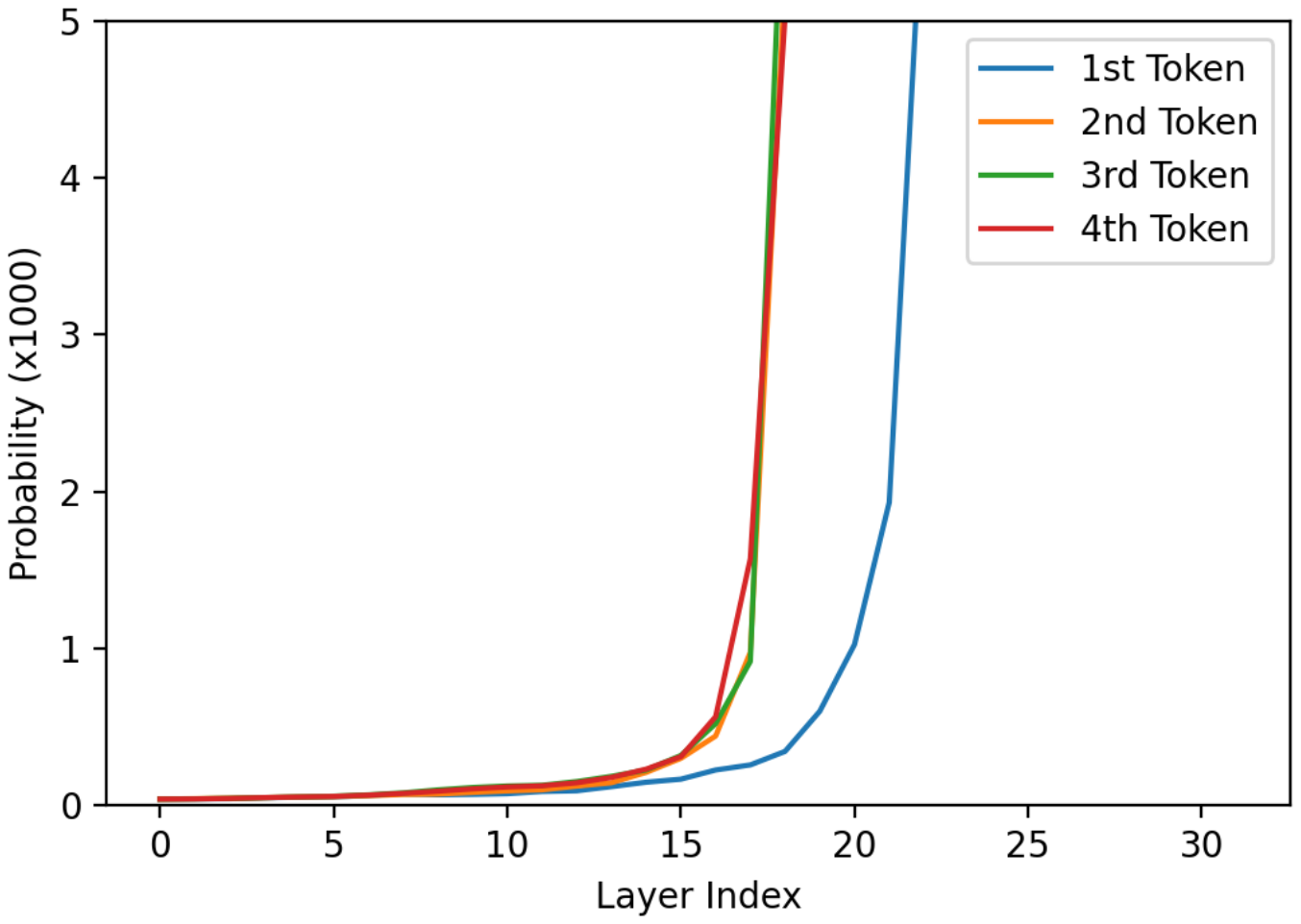}
    \caption{Natural Question}
    \label{fig4:sub2}
  \end{subfigure}
  \begin{subfigure}[b]{0.245\textwidth}
     \includegraphics[width=1\linewidth]{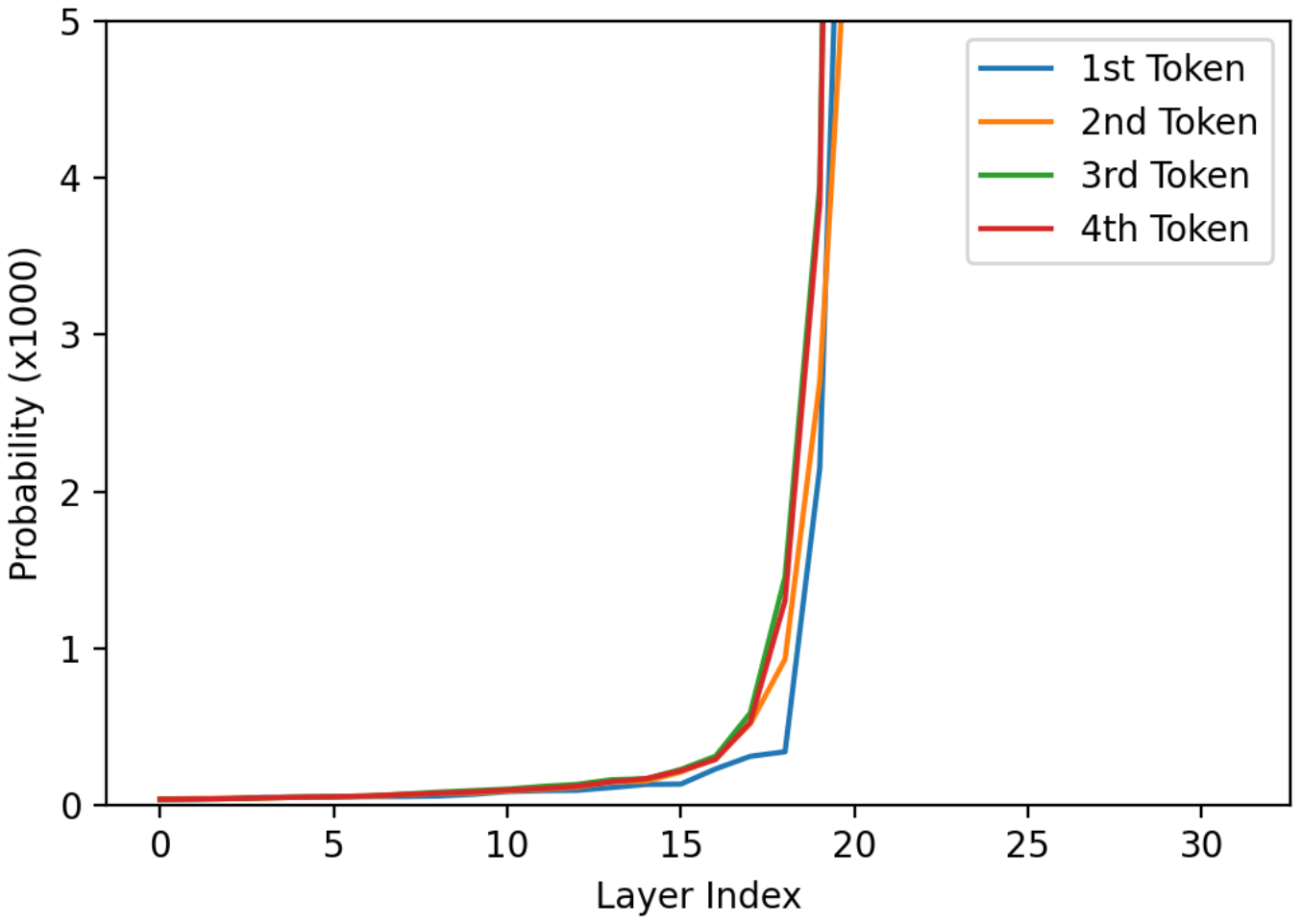}
    \caption{GSM8K}
    \label{fig4:sub3}
  \end{subfigure}
  \begin{subfigure}[b]{0.245\textwidth}
     \includegraphics[width=1\linewidth]{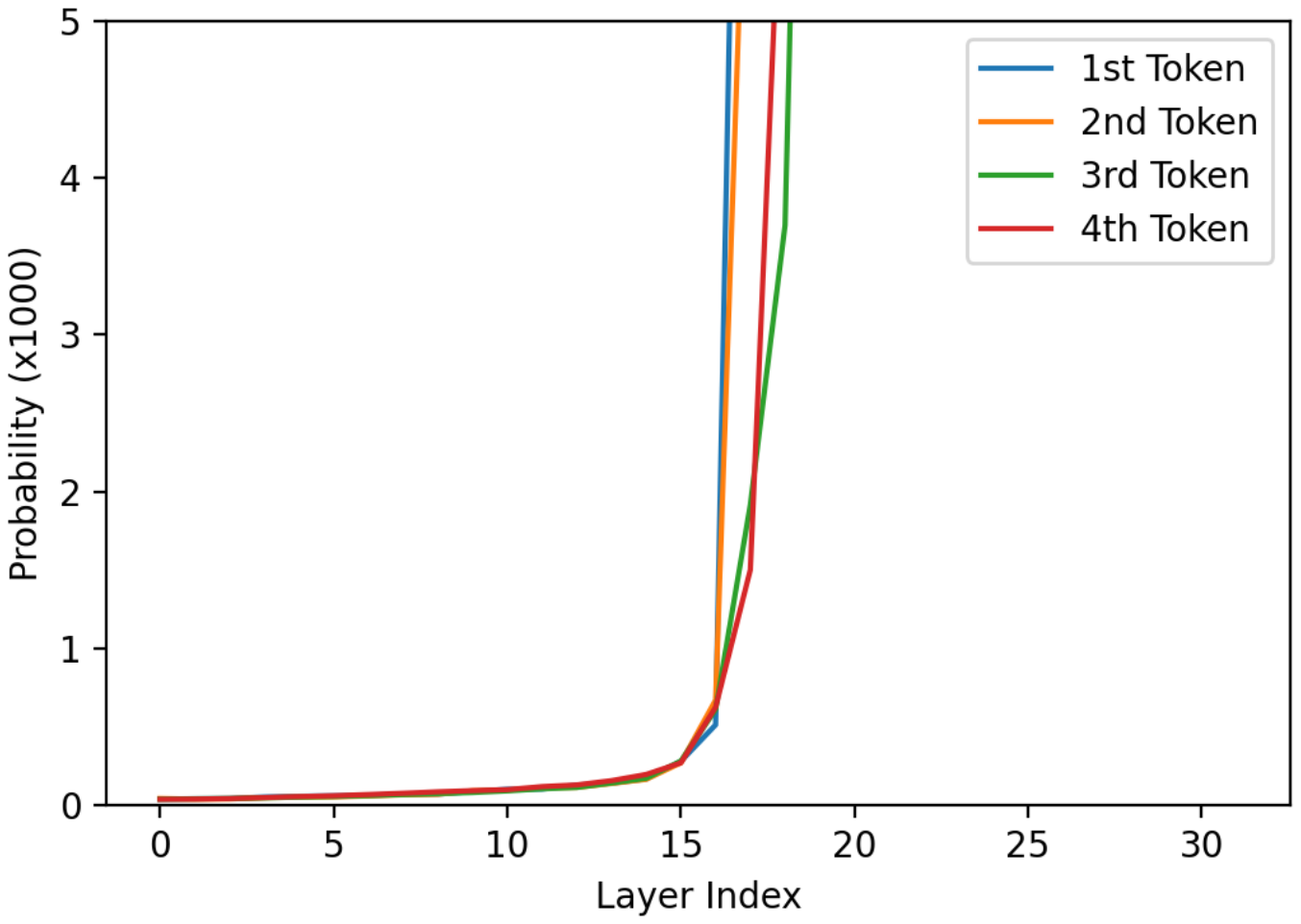}
    \caption{WMT23 (EN->ZH)}
    \label{fig4:sub4}
  \end{subfigure}
    \caption{The average maximum probability of each layer of Llama2-7B-Chat model fine-tuned with LoRA on the four datasets.}
    \label{fig:4}
\end{figure*}

\subsection{GPT-Score} \label{app.2}
For this metric, three powerful LLMs are employed as the evaluators following \citet{chia2024m}. We use the API of GPT-4o \cite{hurst2024gpt}, claude-3-5-sonnet \cite{anthropic2024claude}, and Gemini-1.5-pro \cite{team2024gemini} to generate the responses. The instruction to ask the LLMs to output the scores for the predicted results is shown in Table \ref{table8}. We use the percentage(\%) format of the average scores of these three LLMs as the final scores in our experiments. 

\subsection{Implementation} \label{app.3}
We fine-tuned all the models in our experiments with \textbf{LLaMA-Factory}\footnote{https://github.com/hiyouga/LLaMA-Factory}. We fine-tuned these three models 3 epochs for all datasets. And we set the learning rate to $1e-4$ and LoRA rank to 8. The maximum length of each sample and the batch size are set to 2048 and 16. We sample $M=500$ instances from the validation set to determine the ``boundary layer'', and we set ``boundary layer'' to 15 directly if a dataset does not have a validation set. All the experiments have been completed on one 80G H800 GPU or vGPU-32GB.  


\begin{table*}[t]
  \centering
    \renewcommand\arraystretch{1}
    \scalebox{1}{
    \begin{tabular}{l}
    \toprule
    \multicolumn{1}{c}{Instruction of GPT-Score} \\
    \midrule
    Here is the given dialogue history: \textbf{\{Dialog History\}} \\ 
    Here is the ground truth: \textbf{\{Label Response\}}\\
    Here is the predicted output: \textbf{\{The Response Generated by LLMs\}}\\
    \\ 
     - The aim of this prediction is to generate a summary of the provided historical dialogue. \\
    - Please rate the summary based on its level of abstraction and 
     fluency in summarizing the historical \\ 
     dialogue. The scores should be integers ranging from 0 to 10. \\   - Please give the score number directly without any explanation or introduction.\\
      \bottomrule
    \end{tabular}}
    \caption{The instruction of calculating the GPT-Score in our experiments for \textbf{Samsum} dataset. For translation tasks, we only need to adjust the task description.}
  \label{table8}
\end{table*}

\begin{table*}[t]
  \centering
    \renewcommand\arraystretch{1}
    \scalebox{1}{
    \begin{tabular}{l}
    \toprule
    \multicolumn{1}{c}{Instruction of Evaluating the Generation Quality} \\
    \midrule
    Here is the given dialogue history: \textbf{\{Dialog History\}} \\ 
    Here are two generated responses: \\
    A. \textbf{\{Baseline\}} \\
    B. \textbf{\{Our Method\}} \\
    \\ 
     - The aim of these responses is to generate a summary of the provided historical dialogue. \\
    - Please choose the better response like a human based on four aspects: fluency, accuracy, readability, \\
    and completeness. \\  
    - Please output only ``A'' or ``B'' directly without any explanation or introduction.\\
      \bottomrule
    \end{tabular}}
    \caption{The instruction of evaluating the generation quality for \textbf{Samsum} dataset. ``\textbf{A}'' is the output from baselines and ``\textbf{B}'' is the output from our method. For translation tasks, we also need to adjust the task description.}
  \label{table9}
\end{table*}

\begin{table*}[t]
  \centering
    \renewcommand\arraystretch{0.8}
    \scalebox{1}{
    \begin{tabular}{ll}
    \toprule
    \multicolumn{2}{c}{An Example in Samsum} \\
    \midrule
    \textbf{Eric}: MACHINE! &  \textbf{Rob}: That's so gr8! \\
    \textbf{Eric}: I know! And shows how Americans see Russian &  \textbf{Rob}: And it's really funny! \\
    \textbf{Eric}: I know! I especially like the train part! & \textbf{Rob}: Hahaha! No one talks to the machine \\
    & like that! \\
    \textbf{Eric}: Is this his only stand-up? & \textbf{Rob}: Idk. I'll check. \\ 
    \textbf{Eric}: Sure. & \textbf{Rob}: Turns out no! There are some of his \\
    & stand-ups on youtube. \\
    \textbf{Eric}: Gr8! I'll watch them now! &   \textbf{Rob}: Me too! \\
     \midrule
    \multicolumn{2}{l}{\textbf{Baseline}: Eric and Rob enjoy watching MACHINE stand-up.} \\
     \multicolumn{2}{l}{\textbf{Ours}: Eric and Rob are watching a Russian comedian's stand-up. They will watch more of his videos} \\
     \multicolumn{2}{l}{on YouTube.} \\
      \bottomrule
    \end{tabular}}
    \caption{An example of the test set of Samsum dataset.}
  \label{table10}
\end{table*}

\end{document}